\documentclass{article}
\usepackage{iclr2027_conference,times}

\usepackage{graphicx}
\usepackage{booktabs}
\usepackage{amsmath}
\usepackage{amssymb}
\usepackage{amsthm}
\usepackage{hyperref}
\usepackage{url}
\usepackage{xcolor}
\usepackage{colortbl}
\usepackage{array}
\usepackage{placeins}
\usepackage{needspace}
\definecolor{ablationHeader}{HTML}{F1F5F9}
\definecolor{ablationBlue}{HTML}{EAF5FD}

\title{How Should Teachers Be Prepared?\\
RL on Student-Induced States for On-Policy Distillation}

\author{
\begin{minipage}[t]{\dimexpr\textwidth-2\tabcolsep\relax}
\centering\normalfont
\textbf{
Xiaoyu Ma\textsuperscript{$\spadesuit$}\quad
Haoyue Liu\textsuperscript{$\spadesuit$}\quad
Zhichao Wang\textsuperscript{$\spadesuit$}\quad
Jionghao Zhu\textsuperscript{$\spadesuit$}
}\\[3pt]
\textbf{
Xiaoying Tang\textsuperscript{$\spadesuit\clubsuit\diamondsuit*$}
}\\[5pt]
{\textsuperscript{$\spadesuit$}School of Science and Engineering, The Chinese University of Hong Kong, Shenzhen}\\[-1pt]
{\textsuperscript{$\clubsuit$}Shenzhen Future Network of Intelligence Institute (FNiI-Shenzhen)}\\[-1pt]
{\textsuperscript{$\diamondsuit$}Guangdong Provincial Key Laboratory of Future Networks of Intelligence, CUHK(SZ)}\\[-1pt]
{\textbf{Email:} \texttt{xiaoyuma@link.cuhk.edu.cn, tangxiaoying@cuhk.edu.cn}}\\[2pt]
{\small\textsuperscript{*}Corresponding author.}
\end{minipage}
}

\hypersetup{pdfauthor={Xiaoyu Ma, Haoyue Liu, Zhichao Wang, Jionghao Zhu, Xiaoying Tang}}
\iclrfinalcopy

\begin{document}

\maketitle
\lhead{Preprint}

\begin{abstract}
On-policy distillation (OPD) improves the reasoning capabilities of small language models through token-level teacher supervision on student-generated trajectories.
Yet can teachers that excel at solving problems independently also guide student reasoning effectively?
Prior work shows that when student prefixes follow reasoning paths that differ from the teacher's own or contain errors, teachers can be less accurate when continuing from these prefixes than when solving problems independently.
To this end, we propose Prep-OPD, which uses reinforcement learning (RL) before distillation to train the teacher to adapt to the student's existing reasoning state and correct course when errors arise.
Training optimizes teacher continuations from fixed student prefixes using final-answer correctness as the reward.
The prepared teacher then trains the student through trajectory guidance and token-level supervision.
We evaluate Prep-OPD on eight mathematical reasoning benchmarks, using Qwen3-4B-Instruct-2507 as the teacher and Qwen3-0.6B and Qwen3-1.7B as students.
With the 4B teacher and 1.7B student, Prep-OPD improves average accuracy over standard OPD and the strongest baseline, Relay-OPD, by 8.28 and 2.30 percentage points, respectively.
Controlled experiments further show that teacher RL conditioned on student-generated prefixes yields higher student accuracy than problem-start teacher RL with and without handoff on Qwen3-1.7B. Reusing the same prepared teacher also improves Qwen3-0.6B.

\end{abstract}

\section{Introduction}
\label{sec:intro}

On-policy distillation (OPD) is a promising approach for transferring
reasoning capabilities from strong language models to smaller
students. The student generates its own
trajectories, while the teacher provides dense, token-level supervision
on student-generated prefixes,
often through reverse KL minimization~\citep{gu2024minillm,agarwal2024gkd,zhou2024distillspec}.
Training on student-generated trajectories helps reduce the distribution
mismatch between the prefixes seen during training and those produced
at inference~\citep{lin-etal-2020-autoregressive}.
OPD can also achieve competitive reasoning performance at substantially
lower training cost than a reinforcement learning (RL)
baseline~\citep{opsd2026,jin2026entropy}.

\begin{figure}[!t]
  \centering
  \includegraphics[width=\linewidth]{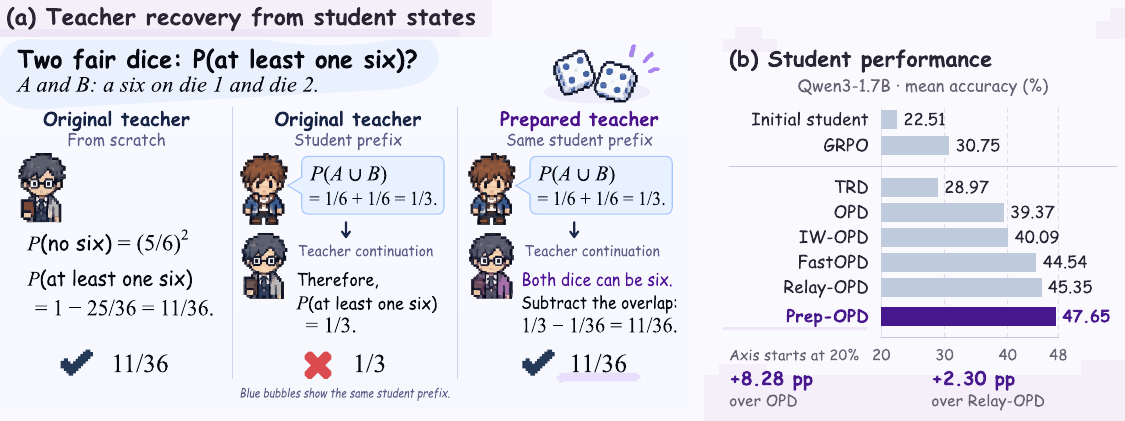}
  \caption{\textbf{Motivation and main results.}
  (a) A solution from scratch and two continuations from the
  same erroneous student prefix.
  (b) Qwen3-1.7B mean benchmark accuracy (\%; Table~\ref{tab:main}).}
  \label{fig:intro-recovery}
\end{figure}

However, OPD implicitly relies on the teacher being a reliable
supervisor on student-generated reasoning states. This assumption can
break down when student prefixes follow reasoning paths different from
those the teacher would take independently, or have already gone in the
wrong direction. In such cases, the teacher can be substantially less
accurate when continuing from these prefixes than when solving the
same problems from scratch. Its token-level feedback
may also be noisy or even misleading~\citep{koo2025switch,fu2026revisiting,xie2026position}.
Thus, strong independent problem-solving ability does not ensure
reliable guidance on student prefixes (Figure~\ref{fig:intro-recovery}(a)).

Existing work improves teacher guidance in several ways.
Veto stabilizes optimization through adaptive distillation targets, but
remains constrained by the teacher's existing continuation
capabilities~\citep{jang2026stable}.
ESR and FastOPD concentrate supervision on early prefixes where teacher
guidance is more effective, but exclude reasoning steps beyond the
truncation point from distillation supervision for that training
instance~\citep{esr2026,fastopd2026}.
TRD improves distillation samples by rewriting complete student
trajectories, whereas SKD and Relay-OPD introduce teacher-generated
tokens or segments during rollouts to intervene in student
trajectories~\citep{trd2026,xu2025skd,relayopd2026}.
However, even with early intervention, teacher continuations may still
be influenced by errors in the student's reasoning.
These methods do not explicitly train the teacher to adapt to the
student's existing reasoning state and correct course when
errors arise.

\begin{figure}[!t]
    \centering
    \includegraphics[width=\linewidth]{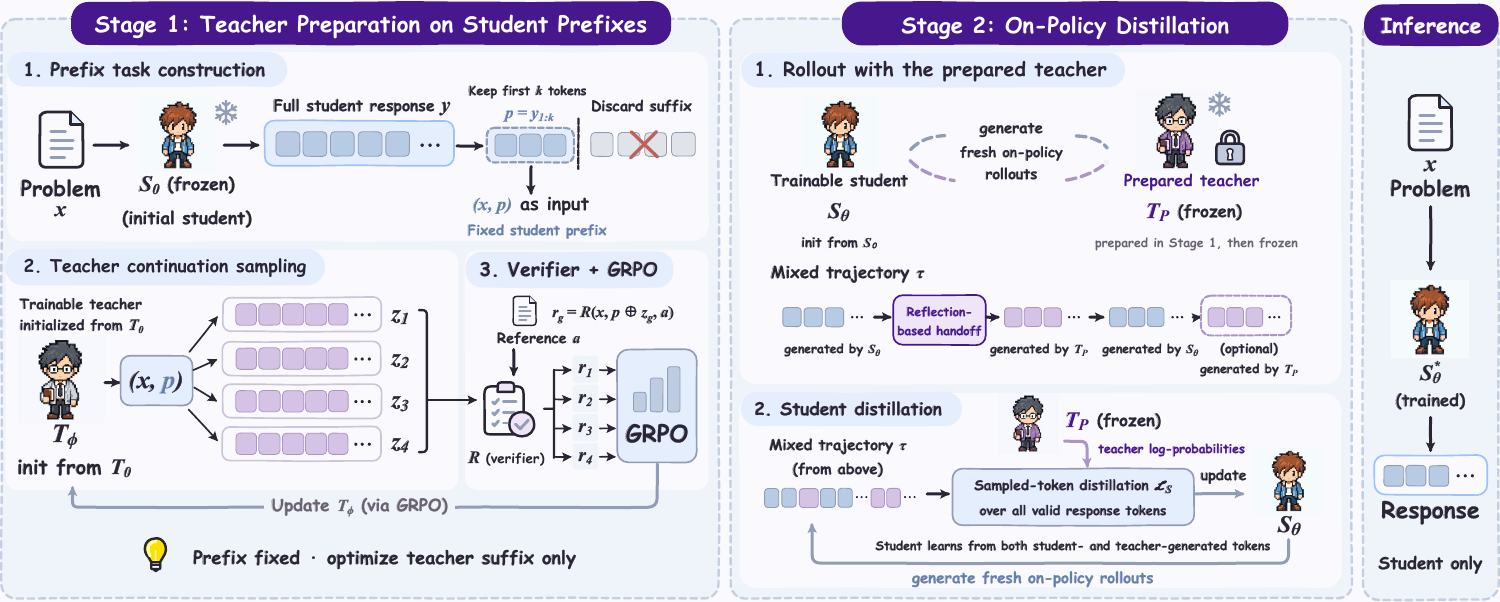}
    \caption{\textbf{Overview of Prep-OPD.}
    Stage 1 fixes prefixes from $S_0$ and trains $T_\phi$, initialized
    from $T_0$, with outcome rewards and suffix-only GRPO. Stage 2 freezes
    $T_P$ for bounded handoffs and token-level supervision of mixed trajectories.
    Blue/purple tokens denote student/teacher generation; distillation
    covers both token types. Inference uses the student alone.}
    \label{fig:recovery-opd}
\end{figure}

This suggests a simple but different strategy: \textbf{prepare the
teacher for student-generated reasoning states before using it
for distillation}.
To this end, we propose Prep-OPD, which uses reinforcement learning
to train the teacher to adapt to the student's existing reasoning
state and correct course when errors arise.
We then distill this ability into the student.
We sample prefixes at different reasoning stages from trajectories
generated by the initial student. The teacher continues reasoning
while preserving each prefix and is trained through reinforcement
learning with final-answer correctness as the reward, yielding the
prepared teacher $T_P$.
We then freeze and reuse the same $T_P$: it supplies bounded
continuations at selected intermediate states in fresh student rollouts
and provides token-level distillation supervision over the entire
trajectory jointly generated by the student and teacher.
The teacher's continuations both provide context for the student's
subsequent reasoning and contribute directly to its updates, allowing
the student to learn to build on existing reasoning and adjust its
direction when needed. Inference uses only the student model
(Figure~\ref{fig:recovery-opd}).

\begin{samepage}
\begin{itemize}
  \item Through controlled continuation experiments, we find that
  \textbf{teachers can be substantially less accurate when continuing
  from student prefixes than when solving independently, and this gap
  persists even at early prefix positions.}
  Problem-start RL improves the teacher's independent problem-solving
  accuracy but provides limited gains in continuation from student
  prefixes. Compared with the problem-start RL teacher, the prepared
  teacher narrows this gap and improves mean continuation accuracy by
  \textbf{2.55 percentage points}.
  \item We propose \textbf{Prep-OPD}, which uses reinforcement learning
  to prepare the teacher before distillation. Conditioned on
  student-generated reasoning prefixes and rewarded for final-answer
  correctness, the teacher learns to adapt to the student's existing
  reasoning state and correct course when errors arise.
  The prepared teacher then distills its learned continuation ability
  into the student through trajectory guidance and token-level supervision.
  \item Under a unified evaluation protocol, we evaluate Prep-OPD on
  eight mathematical reasoning benchmarks with two student sizes.
  With the 4B teacher and 1.7B student, Prep-OPD improves average
  accuracy over standard OPD and the strongest baseline, Relay-OPD,
  by \textbf{8.28 and 2.30 percentage points}, respectively.
\end{itemize}
\end{samepage}
\FloatBarrier

\section{Teacher Supervision on Student Trajectories}
\label{sec:background}

On-policy distillation (OPD) trains a student on trajectories sampled from
its own policy, while a fixed teacher provides token-level supervision
\citep{gu2024minillm,agarwal2024gkd,relayopd2026}. Let $S_\theta$ denote the student,
$S_{\bar\theta}$ its rollout policy, and $T$ the teacher. Given a problem
$x$, the student generates
\begin{equation}
\label{eq:opd-rollout}
    y=(y_1,\ldots,y_N)
    \sim S_{\bar\theta}(\cdot\mid x),
    \qquad
    h_t=(x,y_{<t}).
\end{equation}
Standard OPD typically employs the reverse KL divergence. Taking a
single-sample estimate at the student-sampled token $y_t$, the
teacher--student log-probability difference defines the token-level
advantage
\begin{equation}
\label{eq:opd-advantage}
    \widehat A_t^{\mathrm{OPD}}
    =
    \log T(y_t\mid h_t)
    -
    \log S_{\bar\theta}(y_t\mid h_t).
\end{equation}
A positive advantage increases the probability of the sampled token,
whereas a negative one decreases it. By conditioning supervision on
$h_t$, OPD grounds learning in the reasoning states actually visited by
the student rather than exclusively in teacher-generated trajectories.

However, student-generated reasoning may contain errors or commit to
unproductive directions, which can compromise subsequent teacher
supervision~\citep{fu2026revisiting,xie2026position}. In such cases, the teacher must provide useful guidance
while inheriting the reasoning already produced by the student, rather
than solving the problem again from a clean starting point. A teacher's
ability to solve problems from scratch therefore does not, by itself,
establish its ability to provide effective supervision under
student-generated reasoning. This motivates explicitly preparing the
teacher to continue toward correct solutions from student-induced
reasoning states.
\section{Prep-OPD}
\label{sec:method}

To address this requirement, we propose \textbf{Prep-OPD}, a two-stage
framework that first uses reinforcement learning to train the teacher to
adapt to the student's existing reasoning state and correct course
when errors arise, and then reuses the same prepared teacher to guide and
supervise student learning. In \textbf{Stage 1}, we condition the teacher
on student-generated reasoning prefixes and optimize only its continuations,
using final-answer correctness as the reward, yielding the prepared teacher
$T_P$ (Sections~\ref{sec:prefix-states}--\ref{sec:recovery-rl}).
In \textbf{Stage 2}, we freeze and reuse the same $T_P$ to guide student
trajectories through bounded continuations and provide token-level
distillation supervision over the entire trajectory jointly generated by
the student and teacher (Section~\ref{sec:distill}).
Figure~\ref{fig:recovery-opd} illustrates the overall framework.

\subsection{Student-Conditioned Continuation Tasks}
\label{sec:prefix-states}

To prepare the teacher to adapt to reasoning states actually generated by
the student, we extract unfinished reasoning prefixes from trajectories
of a fixed initial student $S_0$ to construct student-conditioned
continuation tasks. Let $\mathcal{D}$ denote a dataset of
problems $x$ paired with reference answers $a$. For each problem, we first
sample a student response and then retain an unfinished prefix
corresponding to an intermediate reasoning state:
\begin{equation}
\label{eq:recovery-task}
    (x,a)\sim\mathcal{D},
    \qquad
    y\sim S_0(\cdot\mid x),
    \qquad
    k\sim\kappa(\cdot\mid y),
    \qquad
    p=y_{1:k},
\end{equation}
where $\kappa$ is a cut-point distribution over valid intermediate
positions. Repeating this procedure yields a fixed empirical distribution
$q_0$ over continuation tasks $(x,p,a)$.

Each task asks the teacher to continue from the inherited student
reasoning $p$, rather than solve the problem again from a clean starting
point. The reference answer $a$ is not exposed to the teacher and is used
only to evaluate the resulting continuation. Importantly, $p$ remains
fixed throughout the task: the teacher cannot delete or rewrite reasoning
already produced by the student, and any adjustment must therefore be
expressed through subsequent tokens.

Sampling cut points at different depths exposes the teacher to student
reasoning states from different stages of a trajectory, ranging from
relatively early partial solutions to more developed reasoning contexts.

\subsection{Teacher Preparation via Reinforcement Learning}
\label{sec:recovery-rl}

\paragraph{Teacher preparation objective.}
Given continuation tasks sampled from $q_0$, we train the teacher to continue
toward correct solutions while inheriting the student's existing reasoning
$p$. Let $T_0$ denote the initial teacher, and initialize a trainable
teacher $T_\phi$ from $T_0$. Rather than restarting from the original
problem, the teacher continues directly from the inherited student state:
\begin{equation}
\label{eq:recovery-continuation}
    z \sim T_\phi(\cdot \mid x,p),
\end{equation}
where $z$ denotes the teacher-generated continuation. The completed
response is $p\oplus z$, and we optimize
\begin{equation}
\label{eq:recovery-objective}
    J_{\mathrm{prep}}(\phi)
    =
    \mathbb E_{\substack{
        (x,p,a)\sim q_0\\
        z\sim T_\phi(\cdot\mid x,p)
    }}
    \left[
        R(x,p\oplus z,a)
    \right],
\end{equation}
where $R\in\{0,1\}$ evaluates final-answer correctness. This formulation
separates what is inherited from what can still be changed: $p$ fixes the
reasoning already produced by the student, whereas $z$ is the continuation
controlled by the teacher. Final-answer correctness then rewards
continuations that successfully carry the inherited student state toward
a correct solution.

\paragraph{Suffix-only optimization.}
We optimize the above teacher preparation objective using GRPO
\citep{grpo2024}. For each continuation task, the teacher snapshot
$T_{\bar\phi}$ used to generate the current batch samples $G$
continuations $\{z_g\}_{g=1}^{G}$ from the same student prefix. Their rewards
and group-relative advantages are
\begin{equation}
\label{eq:recovery-advantage}
    r_g = R(x,p\oplus z_g,a),
    \qquad
    \widehat A_g
    =
    \frac{r_g-\mu_r}{\sigma_r+\eta},
\end{equation}
where $\mu_r$ and $\sigma_r$ are the mean and standard deviation of
$\{r_g\}_{g=1}^{G}$, and $\eta>0$ is a numerical stabilizer. Since all
continuations in the group share the same problem $x$ and inherited
reasoning $p$, $\widehat A_g$ compares alternative ways of continuing
from the same student-generated state.

For token $t$ in continuation $z_g$, we define
\begin{equation}
\label{eq:teacher-ratio}
    \rho^T_{g,t}(\phi)
    =
    \frac{
        T_\phi(z_{g,t}\mid x,p,z_{g,<t})
    }{
        T_{\bar\phi}(z_{g,t}\mid x,p,z_{g,<t})
    },
\end{equation}
and minimize the clipped surrogate objective
\begin{equation}
\label{eq:recovery-grpo}
    \mathcal L_T(\phi)
    =
    -\widehat{\mathbb E}_{\mathrm{suffix}}
    \left[
        \min\left\{
            \rho^T_{g,t}(\phi)\widehat A_g,\,
            \operatorname{clip}\!\left(
                \rho^T_{g,t}(\phi),
                1-\epsilon_T,
                1+\epsilon_T
            \right)\widehat A_g
        \right\}
    \right].
\end{equation}
Here, $\widehat{\mathbb E}_{\mathrm{suffix}}$ averages over
teacher-generated suffix tokens, with $x$ and $p$ serving only as
conditioning context. Rewards are evaluated on the complete response
$p\oplus z_g$. We keep $q_0$ fixed throughout Stage~1 while sampling
fresh teacher continuations, and denote the resulting frozen prepared
teacher by $T_P$.

\subsection{On-Policy Distillation with the Prepared Teacher}
\label{sec:distill}

Having learned to continue from inherited student reasoning, the frozen
$T_P$ is reused in Stage~2 in two complementary roles: at selected states,
it contributes continuations that steer the student rollout toward more
reliable reasoning directions; over the resulting mixed trajectory, it
provides token-level supervision.

At each handoff, the prefix of the current mixed trajectory plays the same
role as $p$ in Stage~1: it is preserved as inherited reasoning, while $T_P$
determines what comes next. Following Relay-OPD~\citep{relayopd2026},
we use bounded teacher handoffs to insert continuations at selected
prefixes of the mixed trajectory. During a student segment, while the handoff budget
remains, $T_P$ takes over when its most likely next token is a reflection
token~\citep{muennighoff2025s1} and no reflection token appears among
the rollout student $S_{\bar\theta}$'s top-$5$ next-token
candidates at the same prefix.
Equation~\ref{eq:app-handoff} in Appendix~\ref{app:student-config}
specifies this criterion; the same appendix defines the reflection-token
set and handoff budgets. At a handoff, $T_P$ emits the preferred reflection
token and then samples a bounded continuation. After this segment,
control returns to the student unless the rollout has terminated.
The resulting mixed trajectory $\tau=(\tau_1,\ldots,\tau_N)$ contains
both student- and teacher-generated tokens.

The same frozen $T_P$ then provides OPD supervision over the mixed
trajectory. For $h_t^\tau=(x,\tau_{<t})$, we define
\begin{equation}
\label{eq:recovery-opd-advantage}
    \widehat A_t^{P}
    =
    \operatorname{sg}\!\left[
        \log T_P(\tau_t\mid h_t^\tau)
        -
        \log S_{\bar\theta}(\tau_t\mid h_t^\tau)
    \right],
\end{equation}
where $\operatorname{sg}$ denotes stop-gradient. We optimize all realized
response tokens with a clipped sampled-token distillation surrogate;
the complete student update, token
averaging, and rollout budgets are specified in
Appendix~\ref{app:recovery-implementation}.

Teacher-generated continuations therefore both shape the contexts from
which the student resumes generation and participate directly in the
student update, while the same $T_P$ provides dense token-level
supervision over the full mixed trajectory. After distillation, $T_P$
is removed and inference uses the student alone.

% Experimental setup, main results and efficiency, effects of teacher
% continuation, and ablations of teacher preparation, handoff, and roles.
\section{Experiments}
\label{sec:experiments}

\subsection{Experimental Setup}
\label{sec:setup}

\paragraph{Models and datasets.}
We use Qwen3-4B-Instruct-2507 as the initial teacher $T_0$ and
Qwen3-1.7B as the primary student $S_0$, with an additional
Qwen3-0.6B setting \citep{qwen3}. Training problems come from the English subset of
DAPO-Math-17K \citep{dapo2025}. Both student distillation settings use
$16{,}164$ problems, while the GRPO stage of teacher preparation
uses a $2{,}048$-problem training subset.
We evaluate on the eight mathematical reasoning benchmarks used by
\citet{relayopd2026}: AIME 2024, AIME 2025, AIME 2026, AMC 2023,
HMMT February 2026, HMMT November 2025, OlympiadBench, and MATH-500.

\paragraph{Baselines.}
We include the initial student as a reference and compare with
GRPO \citep{grpo2024}, Standard OPD \citep{gu2024minillm,agarwal2024gkd},
TRD \citep{trd2026}, IW-OPD \citep{xie2026position}, FastOPD \citep{fastopd2026}, and
Relay-OPD \citep{relayopd2026}. In the main Qwen3-1.7B comparison, the distillation baselines use
$T_B$, obtained by problem-start RL from $T_0$, whereas Prep-OPD
uses $T_P$, trained on prefixes from $S_0$. The two RL-trained teachers use
the same training-problem pool and outcome verifier. For Qwen3-0.6B, TRD uses
$T_0$, while the other distillation baselines in Table~\ref{tab:main}
use $T_B$. Relay-OPD comparisons using $T_0$ at both student sizes are
reported separately in Appendix~\ref{app:original-teacher-baseline}.

\paragraph{Implementation details.}
We implement training with verl and vLLM on H20 GPUs, with thinking
disabled for both students.
Teacher preparation runs for $128$ GRPO steps; student distillation at
both sizes uses one epoch. We use a $16{,}384$-token generation cap throughout
training and evaluation; FastOPD truncates its training rollouts at
$4{,}096$ tokens. Training and evaluation use temperature $1.0$
and top-$p$ $1.0$. For each training configuration, we select the student
checkpoint with the highest accuracy on a development set.
We evaluate the selected checkpoint using $32$ responses per problem on
the six smaller benchmarks and $4$ on OlympiadBench and MATH-500.
We report mean sampled-response accuracy and its unweighted mean across
the eight benchmarks, giving each benchmark equal weight regardless of size.
For the main results, we compute all eight-benchmark means and
between-method differences before rounding.
Training hyperparameters are summarized in Table~\ref{tab:training-hyperparameters};
further implementation and evaluation details are in Appendix~\ref{app:impl}.

\Needspace{0.48\textheight}
\subsection{Main Results}
\label{sec:main}
\label{sec:student}

% Current Table 1 source: tables/data/main-results.csv.
% Display names map Recovery-OPD/TR to Prep-OPD/TP.
\begin{table}[!t]
  \setlength{\abovecaptionskip}{0pt}
  \caption{\textbf{Mathematical reasoning accuracy} (\%). Avg. is the
  unweighted mean over eight benchmarks. Distillation runs for both the
  $1.7$B and $0.6$B students use $16{,}164$ training problems.
  Bold/underline indicate the best/second-best reported scores
  within each student block.}
  \label{tab:main}
  \label{tab:pairs-main}
  \centering
  \begingroup
  \definecolor{resultblue}{HTML}{ECF7FD}
  \definecolor{modelband}{HTML}{F1F4F7}
  \fontsize{9}{10}\selectfont
  \setlength{\parskip}{0pt}
  \setlength{\tabcolsep}{3pt}
  \setlength{\aboverulesep}{1.5pt}
  \setlength{\belowrulesep}{1.5pt}
  \renewcommand{\arraystretch}{1.04}
  \resizebox{\textwidth}{!}{%
  \begin{tabular}{lcccccccccc}
    \toprule
    \textbf{Method} & \textbf{Teacher} &
    \multicolumn{3}{c}{\textbf{AIME}} & \textbf{AMC} &
    \multicolumn{2}{c}{\textbf{HMMT}} & \textbf{Olympiad} &
    \textbf{MATH} & \textbf{Avg.} \\
    \cmidrule(lr){3-5}\cmidrule(lr){7-8}
    & & 2024 & 2025 & 2026 & 2023 & Feb.~26 & Nov.~25 & Bench & 500 & \\
    \midrule
    \rowcolor{modelband}
    \multicolumn{11}{c}{\textit{Qwen3-0.6B Student}} \\
    Initial student ($S_0$) & -- & 2.08 & 1.25 & 0.00 & 17.19 & 0.00 & 1.25 & 14.15 & 38.35 & 9.28 \\
    GRPO & -- & 9.17 & 13.23 & 12.08 & 46.56 & 6.25 & 7.08 & 35.15 & 65.60 & 24.39 \\
    OPD & $T_B$ & 13.33 & 12.50 & \underline{14.17} & 51.25 & 8.33 & \textbf{9.17} & 39.93 & 70.60 & 27.41 \\
    TRD & $T_0$ & 4.58 & 7.92 & 5.94 & 39.61 & 6.72 & 4.69 & 29.63 & 61.60 & 20.09 \\
    IW-OPD & $T_B$ & \underline{15.00} & 13.33 & \underline{14.17} & \underline{53.75} & 6.82 & 5.83 & 40.30 & 69.20 & 27.30 \\
    FastOPD & $T_B$ & 13.33 & \textbf{21.67} & \underline{14.17} & 53.12 & \underline{9.09} & \textbf{9.17} & 42.52 & 70.80 & 29.23 \\
    Relay-OPD & $T_B$ & \textbf{17.50} & 16.67 & 13.33 & 52.50 & \underline{9.09} & \textbf{9.17} & \underline{45.30} & \underline{73.60} & \underline{29.64} \\
    \rowcolor{resultblue}
    \textbf{Prep-OPD} & $T_P$ & \textbf{17.50} & \underline{18.33} & \textbf{15.00} & \textbf{57.50} & \textbf{10.61} & \underline{7.50} & \textbf{45.48} & \textbf{74.45} & \textbf{30.80} \\
    \midrule
    \rowcolor{modelband}
    \multicolumn{11}{c}{\textit{Qwen3-1.7B Student}} \\
    Initial student ($S_0$) & -- & 9.58 & 8.54 & 5.42 & 42.97 & 4.92 & 3.12 & 37.15 & 68.40 & 22.51 \\
    GRPO & -- & 18.02 & 16.77 & 15.62 & 52.50 & 12.97 & 8.02 & 45.70 & 76.40 & 30.75 \\
    OPD & $T_B$ & 30.00 & 22.50 & 28.33 & 70.00 & 21.97 & 15.00 & 47.63 & 79.50 & 39.37 \\
    TRD & $T_B$ & 19.17 & 15.83 & 13.75 & 48.44 & 11.36 & 4.17 & 44.48 & 74.60 & 28.97 \\
    IW-OPD & $T_B$ & 29.17 & 23.33 & 27.50 & \textbf{77.50} & 16.67 & 17.50 & 49.48 & 79.60 & 40.09 \\
    FastOPD & $T_B$ & \underline{42.50} & 31.15 & 28.44 & \underline{76.48} & 22.35 & \textbf{21.46} & 53.41 & 80.55 & 44.54 \\
    Relay-OPD & $T_B$ & 38.33 & \underline{31.25} & \underline{30.83} & 74.69 & \underline{23.48} & \underline{20.94} & \underline{58.00} & \underline{85.25} & \underline{45.35} \\
    \rowcolor{resultblue}
    \textbf{Prep-OPD} & $T_P$ & \textbf{48.33} & \textbf{31.67} & \textbf{32.08} & 75.62 & \textbf{25.00} & 20.83 & \textbf{60.44} & \textbf{87.20} & \textbf{47.65} \\
    \bottomrule
  \end{tabular}}
  \endgroup
\end{table}

\Needspace{6\baselineskip}
\noindent\textbf{Overall performance.}
Prep-OPD achieves the highest average accuracy in both student
settings (Table~\ref{tab:main}), reaching $47.65\%$ on Qwen3-1.7B
and $30.80\%$ on Qwen3-0.6B. These scores exceed those of the strongest
baseline in Table~\ref{tab:main}, Relay-OPD($T_B$), by $2.30$ and
$1.15$ percentage points, respectively. Prep-OPD obtains the
best or tied-best scores on six of the eight benchmarks for each
student size.

On Qwen3-1.7B, the average improvements over Standard OPD and
Relay-OPD($T_B$) are $8.28$ and $2.30$ points, respectively.
Prep-OPD outperforms Standard OPD on all eight benchmarks and
Relay-OPD($T_B$) on seven. The largest gains over Standard OPD occur
on AIME 2024 and OlympiadBench, at $18.33$ and $12.81$ points,
respectively.

Appendix~\ref{app:seed-comparison} (Table~\ref{tab:seed-comparison}) compares
Prep-OPD and Relay-OPD($T_B$) across seeds 42, 43, and 44.
Prep-OPD achieves higher average accuracy in all three seeds;
the means of the reported run averages are $47.37\%$ and $45.17\%$,
respectively.

\noindent\textbf{Reusing the prepared teacher across students.}
The $0.6$B experiments reuse the same $T_P$ trained on prefixes
from Qwen3-1.7B. The improvements on Qwen3-0.6B show that the
prepared teacher can also benefit a smaller student whose prefixes
were not used during teacher preparation, supporting teacher reuse
across the two tested student sizes.

\begin{figure}[!htbp]
  \centering
  \includegraphics[width=\textwidth]{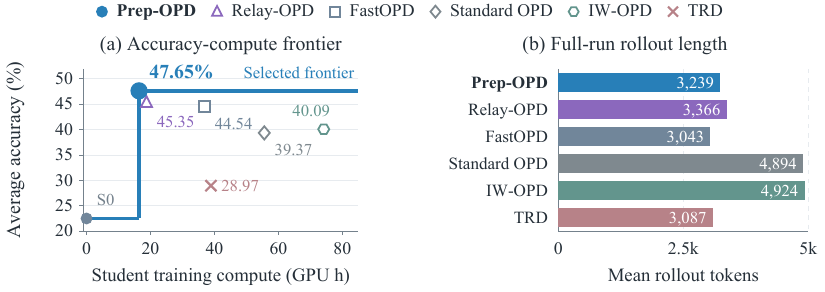}
  \caption{\textbf{Training efficiency on Qwen3-1.7B.}
  (a) Accuracy versus cumulative student training compute for the
  checkpoint selected on the development set for each distillation method.
  Filled markers identify Pareto-optimal points among these selected
  checkpoints; the staircase gives their best accuracy within each compute budget.
  (b) Mean rollout length over the full 126-step student training run
  for each method.}
  \label{fig:training-efficiency}
\end{figure}

\Needspace{4\baselineskip}
\noindent\textbf{Training efficiency.}
Figure~\ref{fig:training-efficiency}(a) compares average benchmark
accuracy against cumulative student training compute on Qwen3-1.7B,
using the checkpoint selected on the development set for each method.
Prep-OPD reaches $47.65\%$ at step 36 with $16.42$ GPU-hours,
compared with $45.35\%$ at $18.75$ GPU-hours for Relay-OPD($T_B$).
This corresponds to $2.30$ percentage points higher accuracy with
$12.4\%$ less student training compute. Relative to the FastOPD
checkpoint in Table~\ref{tab:main} ($44.54\%$ at $36.88$ GPU-hours),
Prep-OPD improves accuracy by $3.11$ points with $55.5\%$ less
student training compute.

Figure~\ref{fig:training-efficiency}(b) reports mean rollout lengths
over the full 126-step student training runs. Prep-OPD averages
$3{,}239$ tokens, compared with $4{,}894$ for Standard OPD and
$4{,}924$ for IW-OPD, reductions of $33.8\%$ and $34.2\%$,
respectively. Its mean length is close to Relay-OPD's $3{,}366$
tokens, while FastOPD ($3{,}043$) and TRD ($3{,}087$) produce
shorter rollouts. FastOPD uses a $4{,}096$-token rollout cap.

\FloatBarrier
\Needspace{0.45\textheight}
\subsection{Teacher Continuation Capability}
\label{sec:teacher}

\begin{figure}[!htbp]
  \centering
  \includegraphics[width=\textwidth]{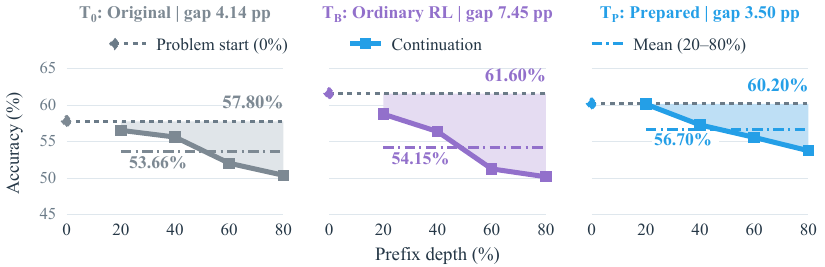}
  \caption{\textbf{Teacher continuation across prefix depths.}
  Accuracy at each depth weights the eight benchmarks equally, and the
  depth mean weights the four continuation positions equally. The $0\%$
  point denotes problem-start solving and is excluded from this mean.
  Panel titles report
  the mean recovery gap in percentage points. Shading shows the pointwise
  gap between problem-start and continuation accuracy, not uncertainty.
  Ordinary RL denotes problem-start RL.}
  \label{fig:conditional}
\end{figure}

This subsection compares problem-start and student-prefix continuation
accuracy for the original teacher $T_0$, the problem-start RL teacher
$T_B$, and the prepared teacher $T_P$ across eight benchmarks
(Figure~\ref{fig:conditional}; Appendix~\ref{app:teacher-conditional}).
All teachers continue the same frozen student prefixes, allowing direct
comparison on shared inherited reasoning states.
Continuation accuracy is averaged equally over prefix depths of $20\%$,
$40\%$, $60\%$, and $80\%$, and overall results weight the eight benchmarks
equally. The \emph{recovery gap}, $G_T=C_T-P_T$, contrasts problem-start
accuracy $C_T$ with mean continuation accuracy $P_T$.

\FloatBarrier

\Needspace{6\baselineskip}
\noindent\textbf{Independent solving leaves a student-state recovery gap.}
$T_0$ achieves a mean accuracy of $57.80\%$ from the problem start and
$53.66\%$ when continuing from student prefixes, a gap of $4.14$
percentage points. This difference reflects the distinction between
producing a solution independently and completing an inherited student
reasoning trajectory.

\Needspace{6\baselineskip}
\noindent\textbf{Gains from problem-start RL do not fully transfer to continuation.}
$T_B$ reaches $61.60\%$ independent accuracy, improving over $T_0$ by
$3.80$ percentage points. Its continuation accuracy is $54.15\%$, an
improvement of only $0.50$ points. The mean gap between problem-start
and continuation accuracy therefore widens from $4.14$ to $7.45$ points.
These results show that gains in independent solving do not yield
corresponding gains in continuation from student prefixes.

\noindent\textbf{Teacher preparation improves continuation accuracy on student prefixes.}
$T_P$ reaches a mean continuation accuracy of $56.70\%$, exceeding
$T_B$ by $2.55$ percentage points. Their independent accuracies are
$60.20\%$ and $61.60\%$, respectively, while the mean recovery gap
narrows from $7.45$ to $3.50$ points. The prepared teacher
achieves higher continuation accuracy despite lower independent accuracy,
indicating that training on student prefixes helps the teacher continue
student reasoning. The benchmark-level results in
Table~\ref{tab:teacher-continuation-matrix} show higher mean continuation
accuracy for $T_P$ than $T_B$ on all eight benchmarks.

\noindent\textbf{Continuation across prefix depths.}
At prefix depths of $20\%$, $40\%$, $60\%$, and $80\%$, $T_P$ exceeds
$T_B$ in mean continuation accuracy by $1.38$, $0.90$, $4.32$, and
$3.58$ percentage points, respectively. Its advantage spans all four
positions, with larger gains at the two deeper positions, where more
student-generated reasoning is already fixed.

\FloatBarrier
\subsection{From Teacher Capability to Student Learning}
\label{sec:ablation}
\label{sec:isolation}

\begin{table}[!htbp]
  \centering
  \caption{\textbf{Ablations of teacher preparation, handoff, and teacher roles.}
  Results are Qwen3-1.7B student accuracies (\%), averaged over the eight
  benchmarks with equal weights.}
  \label{tab:ablation-components}
  \label{tab:factorial}
  \label{tab:roles}
  \begingroup
  \small
  \setlength{\tabcolsep}{3pt}
  \renewcommand{\arraystretch}{1.17}
  \begin{minipage}[t]{0.515\linewidth}
    \centering
    \textbf{(a) Teacher preparation and handoff}\par\vspace{5pt}
    \begin{tabular}{*{4}{>{\centering\arraybackslash}m{\dimexpr.25\linewidth-6pt\relax}}}
      \toprule
      \rowcolor{ablationHeader}
      \rule[-7pt]{0pt}{26pt}Teacher & \shortstack{No\\handoff} & \shortstack{With\\handoff}
        & \shortstack{$\Delta$ handoff\\(pp)} \\
      \midrule
      $T_0$ & 39.96 & 45.33 & $+5.36$ \\
      $T_B$ & 39.37 & 45.35 & $+5.98$ \\
      \rowcolor{ablationBlue}
      $T_P$ & \textbf{41.26} & \textbf{47.65} & $+6.39$ \\
      \bottomrule
    \end{tabular}
  \end{minipage}\hfill
  \begin{minipage}[t]{0.455\linewidth}
    \centering
    \textbf{(b) Generation and supervision}\par\vspace{5pt}
    \begin{tabular}{*{3}{>{\centering\arraybackslash}m{\dimexpr.333333\linewidth-6pt\relax}}}
      \toprule
      \rowcolor{ablationHeader}
      \rule[-7pt]{0pt}{26pt}\shortstack{Generation\\teacher} & \shortstack{Supervision\\teacher}
        & \shortstack{Avg.\\(\%)} \\
      \midrule
      $T_B$ & $T_B$ & 45.35 \\
      $T_P$ & $T_B$ & 46.67 \\
      \rowcolor{ablationBlue}
      $T_P$ & $T_P$ & \textbf{47.65} \\
      \bottomrule
    \end{tabular}
  \end{minipage}
  \par\vspace{4pt}
  \begin{minipage}{\linewidth}
    \small
    No handoff uses Standard OPD; all rows in (b) enable handoff.
    Generation covers handoff decisions and teacher segments;
    supervision scores all response tokens.
    Differences in (a) use unrounded means.
  \end{minipage}
  \endgroup
\end{table}

\paragraph{Student gains with and without teacher handoffs.}
Without teacher handoffs, replacing the problem-start-RL teacher $T_B$
with the prepared teacher $T_P$ increases average student
accuracy from $39.37\%$ to $41.26\%$ ($+1.90$ percentage points;
Table~\ref{tab:ablation-components}(a)). Both configurations use Standard
OPD, where the teacher provides only token-level supervision on
student-generated trajectories. The observed gain therefore does not
require teacher-generated interventions. With handoffs enabled in both
configurations, the same teacher replacement improves accuracy from
$45.35\%$ to $47.65\%$ ($+2.30$ points). These comparisons show that
teacher preparation outperforms problem-start teacher RL in downstream
student performance, both with and without handoffs. Handoffs improve accuracy
for every teacher, while $T_P$ achieves the
highest accuracy in both columns.

\Needspace{8\baselineskip}
\paragraph{Teacher roles in generation and supervision.}
\label{sec:roles}
Each configuration trains its own student online and
induces its own rollout distribution. Comparisons between these configurations
measure conditional effects rather than independent contributions of the two roles.
Holding the supervision teacher fixed at $T_B$, replacing the generation
teacher $T_B$ with $T_P$ increases average student accuracy from
$45.35\%$ to $46.67\%$ (approximately $1.32$ percentage points).
Holding the generation teacher fixed at $T_P$, replacing the supervision
teacher $T_B$ with $T_P$ increases average student accuracy from
$46.67\%$ to $47.65\%$ (approximately $0.98$ percentage points;
Table~\ref{tab:ablation-components}(b)).

\Needspace{8\baselineskip}
\paragraph{Applicability to other distillation methods.}
Using a 4B teacher and a 1.7B student, replacing the standard RL teacher
$T_B$ with the prepared teacher $T_P$ increases average accuracy from
$44.54\%$ to $46.56\%$ for FastOPD and from $40.09\%$ to $41.89\%$
for IW-OPD, yielding gains of $2.02$ and $1.80$ percentage points,
respectively. Both methods achieve higher accuracy on six of the eight
benchmarks (Appendix~\ref{app:other-opd-methods}). These results indicate
that, compared with a standard RL teacher, the prepared teacher can bring
additional gains to different OPD methods.

\FloatBarrier

\section{Related Work}
\label{sec:related}

\paragraph{On-policy distillation.}
On-policy distillation (OPD) builds on knowledge distillation, which transfers teacher distributions or generated sequences to a student \citep{hinton2015distilling,kim2016sequence}. For LLM post-training, OPD evaluates teacher predictions along student-generated rollouts, combining on-policy sampling with dense supervision through token-level divergence losses such as KL \citep{gu2024minillm,agarwal2024gkd}. Related distillation methods refine student-data reuse and contrastive objectives \citep{ko2024distillm,ko2025distillm2}, interpolate teacher--student distributions \citep{shing2025taid}, or use teacher-derived token advantages for preference alignment \citep{gao2025adpa}. By comparison, RL with verifiable rewards (RLVR) often uses sparse binary outcome rewards, which cannot distinguish among failed responses \citep{hubotter2026sdpo}. Standard supervised fine-tuning (SFT) learns from fixed demonstrations rather than trajectories sampled by the current student.

\paragraph{Supervision failure on student prefixes.}
Teacher continuation advantages can diminish as student prefixes lengthen \citep{rethinkingopd2026}. More broadly, teacher hacking shows that closer imitation of an imperfect teacher on fixed offline data can increase divergence from an oracle \citep{tiapkin2025teacherhacking}. Complementary distillation methods select difficult positions and adapt temperatures using teacher--student discrepancy \citep{xie2026adakd}, gate token-level losses through teacher verification of student proposals \citep{huang2026selectkd}, or adjust forward/reverse KL contributions for individual vocabulary entries \citep{jung2025todi}. At the trajectory level, bounded teacher handoffs alter the student's subsequent reasoning states \citep{relayopd2026}.

\paragraph{Teacher adaptation and privileged supervision.}
Distillation with privileged context conditions teacher supervision on reference solutions, demonstrations, or environmental feedback \citep{opsd2026,shenfeld2026sdft,hubotter2026sdpo}. CGD trains students to refine flawed responses using teacher critiques available only during training \citep{kapusuzoglu2026cgd}. Other work jointly optimizes privileged and unprivileged policies with shared parameters \citep{penaloza2026}. SD-Zero first trains a reward-conditioned reviser through supervised fine-tuning and then distills it on policy, whereas PAST applies teacher RL conditioned on complete failed attempts and reference solutions \citep{he2026sdzero,past2026}. Prep-OPD optimizes teacher continuations from fixed intermediate student prefixes using suffix-only GRPO, with reference answers used solely to compute outcome rewards. The prepared teacher is frozen during student training to provide trajectory guidance and token-level supervision.

\section{Conclusion}

We presented Prep-OPD, which uses reinforcement learning to prepare
teachers for continuation from student-generated reasoning prefixes.
The prepared teacher guides student trajectories and provides token-level
supervision, improving student accuracy across eight mathematical reasoning
benchmarks and two student sizes. Our results show that stronger independent
problem-solving ability does not necessarily imply better continuation from
student prefixes. Controlled comparisons further show that student-prefix
preparation improves continuation accuracy and benefits student learning
both with and without handoffs.
Role comparisons support its value in both trajectory generation and
token-level supervision. The same prepared teacher also improves the
smaller student, supporting reuse across the two tested student sizes.

\section*{AI use statement}

We have not used generative AI tools for any of the tasks with required
disclosure, and the remaining required disclosure tasks are not applicable
to this work. Additionally, we used generative AI tools for polishing the
writing and improving the readability of the manuscript.
We also used generative AI tools to assist with literature retrieval and discovery.
We have reviewed
all AI-assisted work: all AI-assisted edits were checked against the
underlying results by the authors. We take responsibility for the final
content of this work, including text, claims or artifacts produced with
the aid of generative AI.

\section*{Reproducibility Statement}
The implementation uses a fixed stack (verl~0.9.0, vLLM~0.21.0) and
a versioned data pipeline: the corpus is deduplicated to $17{,}188$ problems
with a manifest recording every removal, and prefix states carry exact
token IDs with recorded cut indices and exclusion statistics. The main
results use seed 42; additional student runs are reported in
Appendix~\ref{app:seed-comparison}. Teacher and student protocols are described in Section~\ref{sec:setup} and
Appendix~\ref{app:impl}, including checkpoint selection on a development
set and evaluation on the eight-benchmark
suite. Evaluation uses the released benchmark runner
of the base repository. We will release code, configurations, manifests, and the
prefix-state construction tooling in an anonymous repository at submission.

\bibliography{references}
\bibliographystyle{iclr2027_conference}

\appendix
\section{Implementation and Evaluation Details}
\label{app:impl}
\label{app:recovery-implementation}

\subsection{Continuation-task construction and teacher optimization}
\label{app:recovery-construction}

\paragraph{Models and input representation.}
We use the unmodified \texttt{Qwen/Qwen3-1.7B} checkpoint as $S_0$ and
\texttt{Qwen/Qwen3-4B-Instruct-2507} as $T_0$. The same fixed $S_0$
checkpoint supplies the student-generated reasoning prefixes and initializes the trainable
student $S_\theta$; the trainable teacher $T_\phi$ is initialized from
$T_0$. The system message is ``Please reason step by step, and
put your final answer within \texttt{\textbackslash boxed\{\}}.''
The user message contains the problem. We render these messages with the
fixed student tokenizer, using \texttt{add\_generation\_prompt=True}
and \texttt{enable\_thinking=False}. Teacher continuation inputs concatenate
the stored prompt and prefix token IDs directly, continuing the same
unfinished assistant response without re-rendering the prefix or adding
a new assistant turn. The reference answer is stored in a separate
verifier field. Model and tokenizer revisions are fixed across stages,
and rendered input IDs are retained with each task. Token-ID sharing
requires compatible token-to-ID mappings, including the special tokens
used for generation and stopping, across both models.

\paragraph{Prefix construction.}
For each training problem, we sample one response from $S_0$. Let $L_y$
count response tokens preceding the first EOS; if no EOS is emitted,
$L_y$ is the length of the capped response. Eligible cut points satisfy
$\lceil0.15L_y\rceil\le k\le\lfloor0.85L_y\rfloor$ and
$1\le k<L_y$. We exclude prefixes containing a complete, nonempty boxed
answer or a termination token, and states whose prompt and prefix leave
insufficient context for the teacher's continuation budget. From the
remaining positions, we sample up to two uniformly without replacement.
A trajectory with no eligible position contributes no continuation task.
Each selected position defines a separate task $(x,p,a)$ with
$p=y_{1:k}$. The construction does not consult the reference answer.
The retained task records define the fixed empirical distribution $q_0$,
with equal mass assigned to each record. Teacher continuations are sampled
afresh during teacher preparation. We use seed $42$.

\paragraph{Generation budgets.}
We use a common $16{,}384$-token generation cap for initial student
responses, teacher preparation, student training, and benchmark
evaluation. FastOPD truncates its training rollouts at $4{,}096$ tokens.
For teacher preparation, the cap applies to the newly generated suffix
$z$; for handoff training, it covers the full response, counting both
student- and teacher-generated tokens.
Context windows accommodate the conditioning input and the full
generation budget. Problems are retained intact; inputs that leave
insufficient room for the generation budget are excluded. In particular,
the stored prompt and $p$ must leave $16{,}384$ tokens of context capacity
for teacher continuation.
We use temperature $1.0$, top-$p$ $1.0$, unrestricted top-$k$ sampling,
and repetition penalty $1.0$.

\paragraph{Outcome verification.}
The verifier receives the decoded assistant response $p\oplus z$ and
the reference answer $a$ separately; system and user messages are not
scored. We use an ordered grading chain: the first path extracts a boxed
answer and checks it against $a$; the second path uses the supported
\texttt{The answer is:} format; the final path applies symbolic parsing
and equivalence checking with \texttt{math\_verify}. Each path must
establish answer equivalence to yield reward $1$; successful extraction
alone is not sufficient. The reward is $0$ if no path establishes
equivalence or grading raises an exception. The same procedure applies to
EOS-terminated and length-truncated outputs, using the assistant text
actually generated.

\paragraph{Group-relative advantages.}
At each teacher-training iteration, we fix a rollout snapshot
$T_{\bar\phi}$ of the current teacher. For each continuation task
$(x_i,p_i,a_i)$, this snapshot samples $G=8$ continuations
$z_{i,1},\ldots,z_{i,G}$, with rewards
$r_{i,g}=R(x_i,p_i\oplus z_{i,g},a_i)$. Define
\begin{equation}
\label{eq:app-group-statistics}
\begin{aligned}
  \mu_i
  &=\frac{1}{G}\sum_{g=1}^{G}r_{i,g},
  &
  \sigma_i
  &=\sqrt{\frac{1}{G-1}\sum_{g=1}^{G}(r_{i,g}-\mu_i)^2},\\
  \widehat A_{i,g}
  &=\frac{r_{i,g}-\mu_i}{\sigma_i+10^{-6}}.
\end{aligned}
\end{equation}
Here $\sigma_i$ is the sample standard deviation within the common-prefix
group. Continuations from different cut points form separate groups,
even when they originate from the same student response. Zero-variance
groups contribute zero policy advantage and are retained without
additional resampling. We perform no subsequent cross-group or batch
normalization of these advantages.

\paragraph{Teacher optimization.}
At suffix position $t$ of continuation $g$ for task $(x_i,p_i,a_i)$, define
\begin{equation}
\label{eq:app-teacher-ratio}
  \rho^T_{i,g,t}(\phi)
  =\frac{T_\phi(z_{i,g,t}\mid x_i,p_i,z_{i,g,<t})}
  {T_{\bar\phi}(z_{i,g,t}\mid x_i,p_i,z_{i,g,<t})}.
\end{equation}
For both teacher and student updates, write the clipped surrogate as
\begin{equation}
\label{eq:app-clip-surrogate}
  c_\epsilon(\rho,A)
  =\min\!\left\{\rho A,
    \operatorname{clip}(\rho,1-\epsilon,1+\epsilon)A\right\}.
\end{equation}
For a collected teacher batch, let $m^T_{i,g,t}$ mark generated suffix
tokens included in optimization. We minimize
\begin{equation}
\label{eq:app-teacher-loss}
  \mathcal L_T(\phi)
  =-\frac{\sum_{i,g,t}m^T_{i,g,t}\,
  c_{\epsilon_T}\!\left(\rho^T_{i,g,t}(\phi),\widehat A_{i,g}\right)}
  {\sum_{i,g,t}m^T_{i,g,t}}.
\end{equation}
Here $\epsilon_T=0.2$, and the sums cover the global teacher batch across
workers. Each included suffix token has equal weight, matching
$\widehat{\mathbb E}_{\mathrm{suffix}}$ in
Equation~\ref{eq:recovery-grpo}. Prompt, inherited-prefix, and padding
positions have $m^T_{i,g,t}=0$; generated suffix tokens, including genuine
EOS, have $m^T_{i,g,t}=1$. The student prefix conditions the continuation
but contributes no token-level policy-loss term.
Teacher rollout log-probabilities and group advantages remain fixed
while the collected batch is optimized. The rollout snapshot is refreshed
before the next collection. The configured training batch and PPO mini-batch
sizes are both $32$. We use $128$ teacher-training iterations and no
reference-policy KL term.
The final checkpoint is the prepared teacher $T_P$, whose parameters remain
fixed in inference mode throughout student distillation.

\paragraph{Optimization and teacher serving.}
\label{app:hyperparameters}
Table~\ref{tab:training-hyperparameters} summarizes the optimizer, learning
rate, batching, schedule, and method-specific settings.
During student distillation, the teacher is frozen and served with vLLM,
using tensor-parallel size $1$, \texttt{gpu\_memory\_utilization}$=0.45$,
and \texttt{max\_model\_len}$=32{,}768$ tokens. This context limit covers
the conditioning input and generated tokens; the response generation
caps remain as specified above.

\begin{table}[!htbp]
  \centering
  \small
  \setlength{\tabcolsep}{4pt}
  \renewcommand{\arraystretch}{1.15}
  \caption{\textbf{Training hyperparameters.}
  Batch/mini denotes the configured training batch and PPO mini-batch sizes.
  The prepared teacher is denoted by $T_P$ throughout the paper.}
  \label{tab:training-hyperparameters}
  \begin{tabular}{@{}>{\raggedright\arraybackslash}p{0.18\textwidth}cccp{0.105\textwidth}>{\raggedright\arraybackslash}p{0.335\textwidth}@{}}
    \toprule
    Configuration & LR & Optimizer & Batch/mini & Schedule & Additional settings \\
    \midrule
    Teacher RL ($T_B$, $T_P$) & $10^{-6}$ & AdamW & 32/32 & constant & GRPO group size $G=8$; response cap 16,384 tokens \\
    OPD students & $10^{-6}$ & AdamW & 128/128 & constant & One PPO optimization epoch; dynamic batching \\
    Handoff students & $10^{-6}$ & AdamW & 128/128 & constant & Handoff training channel (\texttt{train.sh}) \\
    FastOPD & $10^{-6}$ & AdamW & 128/128 & constant & Training rollout cap 4,096 tokens \\
    IW-OPD & $10^{-6}$ & AdamW & 128/128 & constant & Importance-weight parameter \mbox{$\gamma=0.5$} \\
    \bottomrule
  \end{tabular}
\end{table}

\FloatBarrier

\subsection{Student distillation and baseline configurations}
\label{app:student-config}

\paragraph{Handoff criterion.}
In the implementation, we use the following reflection-disagreement
criterion for Relay-OPD handoffs~\citep{relayopd2026}.
Let $\mathcal R$ be the fixed reflection-token set,
$v_T(h)=\arg\max_v T_P(v\mid h)$, and
$\mathcal K_S(h)=\operatorname{TopK}_K(S_{\bar\theta}(\cdot\mid h))$
the set of the student's top-$K$ next-token IDs at the current mixed
prefix $h$.
During a student segment, while the handoff budget remains, intervention
is triggered when
\begin{equation}
\label{eq:app-handoff}
  g(h)
  =\mathbf{1}[v_T(h)\in\mathcal R]\,
   \mathbf{1}[\mathcal K_S(h)\cap\mathcal R=\varnothing]
  =1.
\end{equation}
We set $K=5$. Both parts of the test use the same prefix $h$.

\paragraph{Reflection-token set.}
The reflection words are \texttt{Wait}, \texttt{But}, \texttt{Hmm},
\texttt{Actually}, \texttt{Hold}, \texttt{However}, \texttt{Yet},
\texttt{Oh}, \texttt{Alternatively}, \texttt{No}, \texttt{Ah},
\texttt{Oops}, and \texttt{Well}. We enumerate the selected tokenizer's
vocabulary, decode each token ID individually, and collect IDs matching
these words, including capitalization and punctuation variants, into
$\mathcal R$. The resulting token-ID set is fixed and reused throughout
Stage~2.

\paragraph{Teacher handoffs.}
Each rollout starts with the student $S_{\bar\theta}$, and permits at
most $M=2$ teacher handoffs. At a trigger, $T_P$ first emits the preferred
reflection token $v_T(h)$ deterministically, then samples a continuation.
Each teacher segment is limited to $256$ generated tokens, including the
initial reflection token, and at most three natural paragraphs. Control
returns to the student as soon as the third paragraph boundary is reached
or the $256$-token cap is reached, whichever occurs first. These limits
apply separately to each teacher segment within the full-response budget.

\paragraph{Scoring and update timing.}
Before collecting a student batch, we fix the rollout snapshot
$S_{\bar\theta}$; $T_P$ remains fixed throughout Stage~2. Once the mixed
trajectories are collected, all actual emitted response tokens are scored
under both $S_{\bar\theta}$ and $T_P$ on their actual mixed prefixes,
before any student parameter update, using full-vocabulary
log-probabilities at temperature $1.0$. For trajectory $i$, let
$h_{i,t}^\tau=(x_i,\tau_{i,<t})$. The saved advantage is
\begin{equation}
\label{eq:app-student-advantage}
  \widehat A_{i,t}^{P}
  =\operatorname{sg}\!\left[
    \log T_P(\tau_{i,t}\mid h_{i,t}^\tau)
    -\log S_{\bar\theta}(\tau_{i,t}\mid h_{i,t}^\tau)
  \right].
\end{equation}
The saved scores include teacher-generated tokens, including the
reflection token emitted at a handoff, scored the same way as other
response tokens. There is no additional group-relative or batch
normalization of $\widehat A_{i,t}^{P}$. Scores and advantages remain
fixed for one student optimization epoch on the collected batch, after
which the rollout snapshot is refreshed. Only the student parameters are
updated.

\paragraph{Student token averaging.}
For problem--trajectory pairs in a collected batch $\mathcal B$, let
$m_{i,t}$ mark included response tokens. Both student-generated and
teacher-generated response tokens, including genuine EOS and each
handoff reflection token, have $m_{i,t}=1$. Prompt, padding, and artificial
control-token positions have $m_{i,t}=0$. The empirical average is
\begin{equation}
\label{eq:app-token-average}
  \widehat{\mathbb E}_{\mathcal B}[f]
  =\frac{\sum_i\sum_t m_{i,t}f_{i,t}}
  {\sum_i\sum_t m_{i,t}}.
\end{equation}
The sums cover the global batch across workers, so each included response
token has equal weight; we do not first normalize each trajectory by its
own length. Tokens retained when a rollout is truncated remain eligible
for optimization. A token added solely to signal forced termination is
excluded rather than treated as a generated EOS.

\paragraph{Student optimization.}
For every included token, define the current-to-rollout student ratio
\begin{equation}
\label{eq:app-student-ratio}
  \rho^S_{i,t}(\theta)
  =\frac{S_\theta(\tau_{i,t}\mid h_{i,t}^\tau)}
  {S_{\bar\theta}(\tau_{i,t}\mid h_{i,t}^\tau)}.
\end{equation}
We apply the clipped sampled-token distillation surrogate of
Relay-OPD~\citep{relayopd2026} with the global token average defined above:
\begin{equation}
\label{eq:app-student-loss}
  \mathcal L_S(\theta)
  =-\widehat{\mathbb E}_{\mathcal B}
  \!\left[
    c_{\epsilon_S}\!\left(
      \rho^S_{i,t}(\theta),\widehat A_{i,t}^{P}
    \right)
  \right],
  \qquad \epsilon_S=0.2.
\end{equation}
The denominator of $\rho^S_{i,t}$ is the stored rollout-student
probability for both student and teacher tokens, so teacher-generated
continuations participate directly in the student loss in addition to
conditioning subsequent student generation. The same frozen $T_P$
supplies the token-level supervision throughout the mixed trajectory.
Student inference uses $S_\theta$ alone.

\label{app:arms}
\begin{table}[h]
  \caption{Full-corpus Qwen3-1.7B distillation configurations ($16{,}164$ student-training
  problems). Budget is the student-training response cap.}
  \label{tab:arms}
  \centering\small
  \setlength{\tabcolsep}{4pt}
  \begin{tabular}{lllrl}
    \toprule
    Arm & Teacher & Channel & Budget & Role \\
    \midrule
    $S_0$ & -- & -- & -- & untrained floor \\
    OPD & $T_B$ & standard & 16K & main distilled baseline \\
    TRD & $T_B$ & offline rewrite & 16K & re-run of \citet{trd2026} \\
    IW-OPD & $T_B$ & standard & 16K & importance-weighted OPD \\
    FastOPD & $T_B$ & truncated rollouts & 4K & re-run of \citet{fastopd2026} \\
    Relay-OPD & $T_B$ & relay & 16K & equal-RL handoff baseline \\
    Prep-OPD & $T_P$ & relay & 16K & full method \\
    \midrule
    \multicolumn{5}{l}{\emph{Additional configurations in Table~\ref{tab:ablation-components}(a,b)}}\\
    OPD ($T_0$) & $T_0$ & standard & 16K & floor cell \\
    Relay ($T_0$) & $T_0$ & relay & 16K & channel-only cell \\
    OPD ($T_P$) & $T_P$ & standard & 16K & teacher-only cell (no handoff) \\
    Cross ($T_P$/$T_B$) & $T_P$ generates, $T_B$ scores & relay & 16K & role-separation arm \\
    \bottomrule
  \end{tabular}
\end{table}

\subsection{Evaluation and checkpoint selection}
\label{app:eval-protocol}
\paragraph{Student training budget.}
Both Qwen3-1.7B and Qwen3-0.6B distillation runs train for one epoch on
$16{,}164$ problems, with a configured training batch size of $128$.
This dataset-level epoch is distinct from the single PPO optimization
epoch applied to each collected batch in
Table~\ref{tab:training-hyperparameters}.

\paragraph{Checkpoint selection.}
We use a fixed development set of $256$ problems for checkpoint selection.
These problems are excluded from training.
For each evaluated checkpoint, we independently sample four responses per
problem, average correctness within each problem, and then average over
all $256$ problems.
For each student-training configuration, we evaluate eight candidate
checkpoints on this same development set and select the checkpoint with
the highest development accuracy. The selected checkpoint is then
evaluated on the eight benchmarks in Section~\ref{sec:setup}; all eight
reported benchmark scores come from that checkpoint.

\begin{figure}[htbp]
  \centering
  \includegraphics[width=\linewidth]{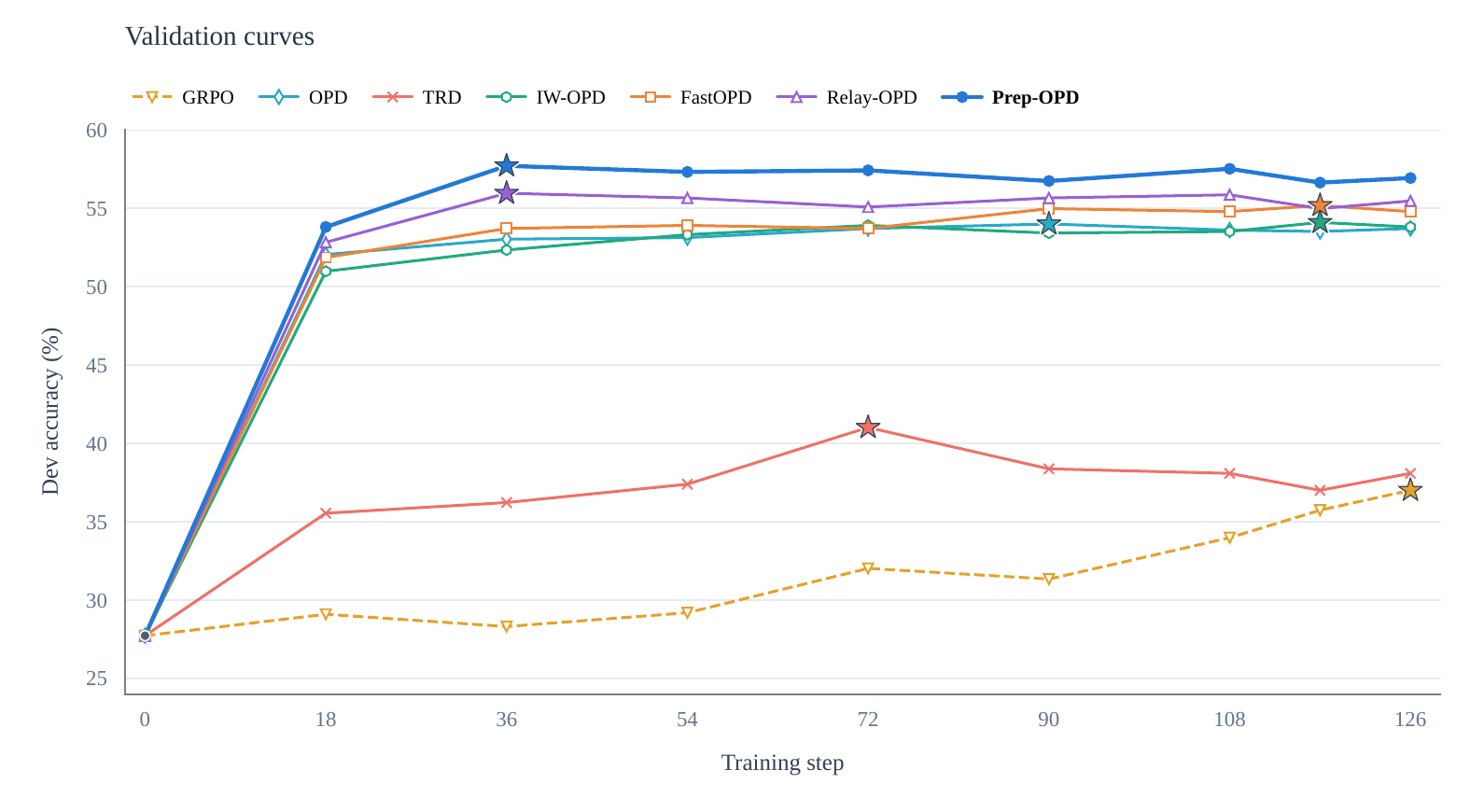}
  \caption{\textbf{Validation curves for Qwen3-1.7B.}
  Development-set accuracy versus training step. Step 0 denotes the
  initial-model reference and is not counted among the eight candidate
  checkpoints. Stars mark the highest plotted development accuracy for
  each method.}
  \label{fig:validation-curves}
\end{figure}

For benchmark evaluation, we sample $32$ responses per problem on the
smaller benchmarks (AIME 2024, AIME 2025, AIME 2026, AMC 2023,
HMMT February 2026, and HMMT November 2025), and $4$ responses per
problem on the larger benchmarks (OlympiadBench and MATH-500).
These sampling counts apply to all evaluated student configurations.
We report mean sampled-response accuracy for each benchmark and the
unweighted mean across the eight benchmarks.

\paragraph{Implementation checks.}
\label{app:tests}
Five acceptance tests were run before any training arm, all passing:
exact-prefix token identity between construction and continuation (32/32
static and 32/32 clean-continuation assertions); suffix-mask correctness
on real pipeline tensors; the reward parser (12 assertions and 11
observational cases covering multi-boxed answers, prefix-wrong-suffix-correct
relay shapes, unit and format variants, and non-crashing behavior on
empty/noisy inputs); teacher isolation (checkpoints contain student-actor
weights only); and checkpoint reload parity (all 310 floating-point tensors
within $3.8\times10^{-6}$ of the base, consistent with two AdamW steps at
lr $10^{-6}$).

\subsection{Teacher conditional-capability evaluation}
\label{app:teacher-conditional}
Figure~\ref{fig:conditional} and Table~\ref{tab:teacher-continuation-matrix}
report teacher problem-start and student-prefix continuation accuracy
across the eight benchmarks in Section~\ref{sec:setup}.
The two RL teachers share the same initialization,
training problem pool, outcome verifier, and update budget; their training
condition is problem-start solving for $T_B$ and student-prefix continuation
for $T_P$. We construct a single evaluation manifest with seed $42$ from
fixed $S_0$ outputs. All three teachers, $T_0$, $T_B$, and $T_P$, share
the same problem set and prefix collection. They correspond to
\texttt{t0}, \texttt{tb}, and \texttt{tr}, respectively, in the evaluation
configuration. The evaluation conditions are
problem-start solving (clean, or $0\%$) and continuation from prefixes at
$20\%$, $40\%$, $60\%$, and $80\%$ of the fixed student response's token
length.
These fixed evaluation positions differ from the randomized training-prefix
construction in Appendix~\ref{app:recovery-construction}. Prefix validity
checks follow that construction protocol. No teacher-specific triggers or
subsequent correctness outcomes select the states.

Problem-start solving generates a full response from the problem alone;
student-conditioned continuation generates a suffix from the frozen
prefix, with the final answer verified over the complete prefix and
suffix. The two conditions use the same decoder (temperature $1.0$,
top-$p$ $1.0$) and the same generated-token cap. The prefix includes
computation already spent by the student, so the comparison measures
conditional capability rather than equal-total-inference-cost accuracy.
Both conditions for a given teacher use the same final checkpoint.

Continuation samples are first averaged within problem and depth, then
over the evaluated problems within each benchmark. Each benchmark's
continuation accuracy is the equal-weight mean across the four depths;
overall accuracies weight the eight benchmarks equally. Independent
accuracy averages over evaluated problems within each benchmark and then
weights the eight benchmarks equally.

Figure~\ref{fig:conditional} compares the two conditions separately
for $T_0$, $T_B$, and $T_P$, using a common accuracy scale. The $0\%$
point denotes problem-start solving, and a horizontal dashed line marks
its accuracy. A thick curve shows continuation accuracy at $20\%$,
$40\%$, $60\%$, and $80\%$; a colored dash-dot line marks the
equal-weight mean across these four continuation depths, excluding $0\%$. Problem-start
accuracy and the depth mean are labeled directly on their reference lines;
each panel title reports the mean gap in percentage points. Four curve
markers identify the evaluated depths. The shaded region
shows the pointwise gap between the problem-start reference and continuation
curve, not a confidence interval. The area of the interpolated region is not
used to calculate the mean gap. The problem-start accuracies are
$57.80\%$, $61.60\%$, and $60.20\%$, while the continuation means are
$53.66\%$, $54.15\%$, and $56.70\%$, yielding mean gaps of $4.14$,
$7.45$, and $3.50$ percentage points, respectively.

% Generated from figures/data/新种子啊.xlsx by import_teacher_continuation.py.
\begin{table}[p]
  \caption{\textbf{Detailed teacher continuation results.}
  Problem-start accuracy $C_T$ and continuation accuracy at four student-prefix
  depths are reported in percent. $P_T$ is the equal-weight mean over the
  four depths; $G_T=C_T-P_T$ is the recovery gap in percentage points.}
  \label{tab:teacher-continuation-matrix}
  \centering
  \normalsize
  \setlength{\tabcolsep}{4pt}
  \renewcommand{\arraystretch}{1.07}
  \begin{tabular}{@{}llrrrrrrr@{}}
    \toprule
    & & & \multicolumn{4}{c}{Student-prefix depth} & & \\
    \cmidrule(lr){4-7}
    Benchmark & Teacher & $C_T$ & $20\%$ & $40\%$ & $60\%$ & $80\%$ & $P_T$ & $G_T$ \\
    \midrule
    AIME 2024 & $T_0$ & 60.10 & 60.94 & 56.15 & 53.33 & 54.17 & 56.15 & 3.95 \\
     & $T_B$ & 67.92 & 60.42 & 54.79 & 50.10 & 44.90 & 52.55 & 15.37 \\
     & $T_P$ & 59.27 & 61.46 & 59.48 & 53.65 & 56.15 & 57.68 & 1.58 \\
    \addlinespace[3pt]
    AIME 2025 & $T_0$ & 41.04 & 45.31 & 46.46 & 38.23 & 38.13 & 42.03 & -0.99 \\
     & $T_B$ & 53.13 & 44.58 & 41.35 & 35.73 & 38.02 & 39.92 & 13.21 \\
     & $T_P$ & 44.90 & 45.10 & 42.81 & 44.58 & 41.88 & 43.59 & 1.31 \\
    \addlinespace[3pt]
    AIME 2026 & $T_0$ & 48.23 & 40.31 & 43.54 & 38.85 & 31.67 & 38.59 & 9.64 \\
     & $T_B$ & 52.29 & 51.88 & 42.50 & 39.38 & 38.02 & 42.94 & 9.34 \\
     & $T_P$ & 55.83 & 53.23 & 45.52 & 46.56 & 34.90 & 45.05 & 10.78 \\
    \addlinespace[3pt]
    AMC 2023 & $T_0$ & 89.38 & 81.80 & 79.84 & 83.28 & 83.05 & 81.99 & 7.39 \\
     & $T_B$ & 93.67 & 90.00 & 84.38 & 81.72 & 82.34 & 84.61 & 9.06 \\
     & $T_P$ & 90.70 & 89.14 & 87.27 & 87.81 & 83.20 & 86.86 & 3.84 \\
    \addlinespace[3pt]
    HMMT Feb. 2026 & $T_0$ & 28.88 & 31.53 & 26.61 & 26.80 & 23.67 & 27.15 & 1.73 \\
     & $T_B$ & 30.97 & 28.31 & 31.44 & 26.14 & 25.66 & 27.89 & 3.08 \\
     & $T_P$ & 28.88 & 31.72 & 29.55 & 27.37 & 29.55 & 29.55 & -0.67 \\
    \addlinespace[3pt]
    HMMT Nov. 2025 & $T_0$ & 41.25 & 41.35 & 39.58 & 31.04 & 28.23 & 35.05 & 6.20 \\
     & $T_B$ & 43.75 & 40.42 & 41.25 & 31.98 & 32.71 & 36.59 & 7.16 \\
     & $T_P$ & 43.54 & 41.56 & 39.06 & 34.48 & 32.40 & 36.88 & 6.66 \\
    \addlinespace[3pt]
    OlympiadBench & $T_0$ & 64.30 & 65.26 & 66.56 & 63.11 & 64.52 & 64.86 & -0.56 \\
     & $T_B$ & 65.48 & 64.67 & 66.41 & 60.67 & 60.26 & 63.00 & 2.48 \\
     & $T_P$ & 68.81 & 64.93 & 67.07 & 65.56 & 66.52 & 66.02 & 2.79 \\
    \addlinespace[3pt]
    MATH-500 & $T_0$ & 89.20 & 85.95 & 86.30 & 81.70 & 79.70 & 83.41 & 5.79 \\
     & $T_B$ & 85.60 & 89.90 & 89.00 & 84.40 & 79.55 & 85.71 & -0.11 \\
     & $T_P$ & 89.65 & 94.05 & 87.60 & 84.70 & 85.50 & 87.96 & 1.69 \\
    \midrule
    Macro average & $T_0$ & 57.80 & 56.56 & 55.63 & 52.04 & 50.39 & 53.66 & 4.14 \\
     & $T_B$ & 61.60 & 58.77 & 56.39 & 51.26 & 50.18 & 54.15 & 7.45 \\
     & $T_P$ & 60.20 & 60.15 & 57.30 & 55.59 & 53.76 & 56.70 & 3.50 \\
    \bottomrule
  \end{tabular}
  \par\smallskip
  \begin{minipage}{\textwidth}
    \footnotesize
    Macro averages weight the eight benchmarks equally. All means and gaps
    are computed before rounding to two decimal places.
    $T_0$: original teacher; $T_B$: problem-start RL teacher;
    $T_P$: prepared teacher.
  \end{minipage}
\end{table}

\FloatBarrier

\subsection{Compute accounting}
\label{app:compute}
Figure~\ref{fig:training-efficiency}(a) reports cumulative student
training compute through each selected checkpoint, derived from logged
student-step times and an allocation of eight GPUs. Prefix generation
for teacher preparation and teacher RL are separate stages and are
excluded from this student-training measure.
The selected Prep-OPD, Relay-OPD, and FastOPD checkpoints use
$16.42$, $18.75$, and $36.88$ GPU-hours, respectively.
Figure~\ref{fig:training-efficiency}(b) averages rollout length over
the full 126-step student training runs, with means of $3{,}239$,
$3{,}366$, $3{,}043$,
$4{,}894$, $4{,}924$, and $3{,}087$ tokens for Prep-OPD,
Relay-OPD, FastOPD, Standard OPD, IW-OPD, and TRD, respectively.

The mixed-teacher configuration in Table~\ref{tab:ablation-components}(b)
deploys $T_P$
for generation and $T_B$ for scoring, so its per-step cost includes both
teachers' inference -- the handoff engine serving $T_P$ at handoff
points and a second forward pass through $T_B$ to score every realized
token. This configuration examines conditional changes in the generation
and scoring roles.

\Needspace{10\baselineskip}
\subsection{Comparison with the original-teacher baseline}
\label{app:original-teacher-baseline}
Relay-OPD with $T_B$ achieves the highest average accuracy among the
baselines in Table~\ref{tab:main} at both student sizes.
Table~\ref{tab:original-teacher-baseline} compares Prep-OPD with the
corresponding Relay-OPD variant using the original teacher $T_0$.
The Prep-OPD results at both student sizes reproduce the corresponding
rows of Table~\ref{tab:main}. The $0.6$B Relay-OPD ($T_0$) results are
reported here as an additional original-teacher comparison;
the $1.7$B Relay-OPD ($T_0$) results correspond to the original-teacher
configuration with handoff in Table~\ref{tab:ablation-components}(a).
Prep-OPD uses $T_P$ throughout this comparison.

For Qwen3-0.6B, Prep-OPD improves average accuracy from $29.02\%$ to
$30.80\%$, a gain of $1.78$ percentage points, and outperforms
Relay-OPD ($T_0$) on six of eight benchmarks. For Qwen3-1.7B, average
accuracy increases from $45.33\%$ to $47.65\%$, a gain of $2.32$ points
computed before rounding, with improvements on seven of eight benchmarks.
Prep-OPD therefore retains an average-accuracy advantage over the
original-teacher version of Relay-OPD at both student sizes.

% Source: tables/data/original-teacher-baseline.csv; component inputs remain unchanged.
% Recomputed from the current packaged CSVs, including the authorized entry corrections.
\begin{table}[!htbp]
  \caption{\textbf{Comparison with Relay-OPD using the original teacher.}
  Accuracy is reported in percent; $\Delta$ is Prep-OPD minus Relay-OPD
  in percentage points. Bold marks the higher accuracy within each student
  size. Avg. weights the eight benchmarks equally.
  Differences are computed before rounding.}
  \label{tab:original-teacher-baseline}
  \centering\small
  \setlength{\tabcolsep}{4pt}
  \renewcommand{\arraystretch}{1.08}
  \begin{tabular}{@{}lrrrrrr@{}}
    \toprule
    & \multicolumn{3}{c}{Qwen3-0.6B} & \multicolumn{3}{c}{Qwen3-1.7B} \\
    \cmidrule(lr){2-4}\cmidrule(lr){5-7}
    Benchmark & Relay ($T_0$) & Prep ($T_P$) & $\Delta$
              & Relay ($T_0$) & Prep ($T_P$) & $\Delta$ \\
    \midrule
    AIME 2024 & 13.33 & \textbf{17.50} & $+4.17$ & 43.65 & \textbf{48.33} & $+4.69$ \\
    AIME 2025 & 14.17 & \textbf{18.33} & $+4.17$ & 28.33 & \textbf{31.67} & $+3.33$ \\
    AIME 2026 & \textbf{18.33} & 15.00 & $-3.33$ & 31.15 & \textbf{32.08} & $+0.94$ \\
    AMC 2023 & 50.62 & \textbf{57.50} & $+6.88$ & 72.50 & \textbf{75.62} & $+3.12$ \\
    HMMT Feb. 2026 & 8.33 & \textbf{10.61} & $+2.27$ & 23.77 & \textbf{25.00} & $+1.23$ \\
    HMMT Nov. 2025 & \textbf{10.83} & 7.50 & $-3.33$ & \textbf{21.67} & 20.83 & $-0.83$ \\
    OlympiadBench & 44.07 & \textbf{45.48} & $+1.41$ & 57.26 & \textbf{60.44} & $+3.19$ \\
    MATH-500 & 72.45 & \textbf{74.45} & $+2.00$ & 84.30 & \textbf{87.20} & $+2.90$ \\
    \midrule
    Avg. & 29.02 & \textbf{30.80} & $+1.78$ & 45.33 & \textbf{47.65} & $+2.32$ \\
    \bottomrule
  \end{tabular}
\end{table}

\FloatBarrier
\Needspace{16\baselineskip}
\subsection{Comparison across random seeds}
\label{app:seed-comparison}
We compare Prep-OPD with Relay-OPD($T_B$), the strongest baseline
in Table~\ref{tab:main}, on Qwen3-1.7B using seeds 42, 43, and 44.
Table~\ref{tab:seed-comparison} combines the original seed-42 results
with the two additional runs for each method.
Each row reports the eight benchmark accuracies and their unweighted mean.

\begin{table}[!htbp]
  \caption{\textbf{Comparison across random seeds on Qwen3-1.7B.}
  Accuracy is reported in percent. The seed-42 rows reproduce
  Table~\ref{tab:main}; seeds 43 and 44 are additional runs.
  Avg. is the unweighted mean over the eight benchmarks, computed
  before rounding the benchmark entries.}
  \label{tab:seed-comparison}
  \centering
  \begingroup
  \fontsize{9}{10}\selectfont
  \setlength{\tabcolsep}{3pt}
  \renewcommand{\arraystretch}{1.08}
  \resizebox{\textwidth}{!}{%
  \begin{tabular}{llrrrrrrrrr}
    \toprule
    \textbf{Method} & \textbf{Seed} &
    \multicolumn{3}{c}{\textbf{AIME}} & \textbf{AMC} &
    \multicolumn{2}{c}{\textbf{HMMT}} & \textbf{Olympiad} &
    \textbf{MATH} & \textbf{Avg.} \\
    \cmidrule(lr){3-5}\cmidrule(lr){7-8}
    & & 2024 & 2025 & 2026 & 2023 & Feb.~26 & Nov.~25 & Bench & 500 & \\
    \midrule
    Prep-OPD ($T_P$) & 42 & 48.33 & 31.67 & 32.08 & 75.62 & 25.00 & 20.83 & 60.44 & 87.20 & 47.65 \\
    & 43 & 45.83 & 30.83 & 34.17 & 75.00 & 23.48 & 20.83 & 59.93 & 86.30 & 47.05 \\
    & 44 & 43.33 & 31.67 & 37.50 & 73.75 & 26.52 & 23.33 & 58.30 & 85.00 & 47.42 \\
    \midrule
    Relay-OPD ($T_B$) & 42 & 38.33 & 31.25 & 30.83 & 74.69 & 23.48 & 20.94 & 58.00 & 85.25 & 45.35 \\
    & 43 & 40.00 & 29.58 & 29.58 & 74.38 & 23.48 & 20.83 & 56.67 & 84.45 & 44.87 \\
    & 44 & 40.62 & 29.79 & 31.04 & 75.00 & 23.11 & 19.79 & 57.41 & 85.45 & 45.28 \\
    \bottomrule
  \end{tabular}}
  \endgroup
\end{table}

The means of the three reported run-level averages are $47.37\%$
for Prep-OPD and $45.17\%$ for Relay-OPD($T_B$).
The corresponding gains at seeds 42, 43, and 44 are $2.30$, $2.18$,
and $2.14$ percentage points. Averaging these gains before rounding
gives $2.21$ points. Prep-OPD has higher average accuracy in all
three runs.
\FloatBarrier
\Needspace{16\baselineskip}

\subsection{Applicability to other distillation methods}
\label{app:other-opd-methods}
To test whether teacher-preparation training can benefit other OPD
methods, we add comparisons with FastOPD and IW-OPD, using a 4B teacher
and a 1.7B student. Table~\ref{tab:other-opd-methods} reports student
performance with the standard RL teacher $T_B$ and the prepared teacher
$T_P$ on the eight benchmarks.

% Source: rebuttal/rebuttal.md, Reviewer myxo Q1, Table 3.
\begin{table}[!htbp]
  \caption{\textbf{Student performance of FastOPD and IW-OPD with the
  standard RL teacher $T_B$ and the prepared teacher $T_P$.}
  The teacher is 4B and the student is 1.7B. Accuracy is reported in
  percent; Avg. is the unweighted mean across the eight benchmarks.}
  \label{tab:other-opd-methods}
  \centering
  \begingroup
  \fontsize{9}{10}\selectfont
  \setlength{\tabcolsep}{3pt}
  \renewcommand{\arraystretch}{1.08}
  \resizebox{\textwidth}{!}{%
  \begin{tabular}{lrrrrrrrrr}
    \toprule
    \textbf{Method / Teacher} &
    \multicolumn{3}{c}{\textbf{AIME}} & \textbf{AMC} &
    \multicolumn{2}{c}{\textbf{HMMT}} & \textbf{Olympiad} &
    \textbf{MATH} & \textbf{Avg.} \\
    \cmidrule(lr){2-4}\cmidrule(lr){6-7}
    & 2024 & 2025 & 2026 & 2023 & Feb.~26 & Nov.~25 & Bench & 500 & \\
    \midrule
    FastOPD ($T_B$) & 42.50 & 31.15 & 28.44 & 76.48 & 22.35 & 21.46 & 53.41 & 80.55 & 44.54 \\
    FastOPD ($T_P$) & 44.58 & 30.83 & 31.25 & 76.33 & 25.19 & 24.58 & 56.70 & 83.05 & \textbf{46.56 (+2.02)} \\
    \midrule
    IW-OPD ($T_B$) & 29.17 & 23.33 & 27.50 & 77.50 & 16.67 & 17.50 & 49.48 & 79.60 & 40.09 \\
    IW-OPD ($T_P$) & 35.21 & 25.21 & 29.79 & 76.09 & 18.75 & 17.29 & 52.33 & 80.45 & \textbf{41.89 (+1.80)} \\
    \bottomrule
  \end{tabular}}
  \endgroup
\end{table}

\FloatBarrier

\end{document}